# Identifying relationships between drugs and medical conditions: winning experience in the Challenge 2 of the OMOP 2010 Cup

Vladimir Nikulin[#]
Department of Mathematical Methods in Economics, Vyatka State University, Kirov, Russia

## Abstract

There is a growing interest in using a longitudinal observational databases to detect drug safety signal. In this paper we present a novel method, which we used online during the OMOP Cup. We consider homogeneous ensembling, which is based on random re-sampling (known, also, as bagging) as a main innovation compared to the previous publications in the related field. This study is based on a very large simulated database of the 10 million patients records, which was created by the Observational Medical Outcomes Partnership (OMOP). Compared to the traditional classification problem, the given data are unlabelled. The objective of this study is to discover hidden associations between drugs and conditions. The main idea of the approach, which we used during the OMOP Cup is to compare the numbers of observed and expected patterns. This comparison may be organised in several different ways, and the outcomes (base learners) may be quite different as well. It is proposed to construct the final decision function as an ensemble of the base learners. Our method was recognised formally by the Organisers of the OMOP Cup as a top performing method for the Challenge N2.

**Keywords:** longitudinal observational data, signal detection, temporal pattern discovery, unsupervised learning, electronic health records

## INTRODUCTION

An improvement of drug safety and the identification of adverse drug events remains a very important problem. Several recent drug safety events have highlighted the need for new data sources and algorithms to assist in identifying adverse drug events in a more timely, effective, and efficient manner. The methods and statistical tools used on large healthcare data sources (e.g., administrative claims and electronic health records) have been lacking and are not yet systematized to look at disparate databases. The Observational Medical Outcomes Partnership (OMOP) conducted a Cup Competition as a catalyst for new methods development to identify relationships in data between drugs and adverse events or conditions (OMOP Newsletter, 2010).

To provide an objective basis for monitoring and assessing the safety of marketed products, pharmaceutical companies and regulatory agencies have implemented post-marketing surveillance activities based in large measure on the collection of spontaneously generated adverse reaction reports. Report initiation (by health professionals and consumers) is generally voluntary; by contrast, the pharmaceutical companies are generally under legal obligation to follow up on reports that they receive and to pass them along to various regulatory authorities (Fram et al., 2003).

Every drug has undergone extensive testing before being released to the market, but even pre-marketing clinical trials involving thousands of people cannot uncover all adverse events that may occur in a much larger and diverse population. Traditionally, post-marketing safety signal detection has relied on voluntary, spontaneous reporting of suspected adverse drug reactions by health care professionals, patients, and consumers (Schuemie, 2010). There is a global interest in using electronic health records for active drug safety surveillance. Many methods have been developed and exploited for quantitative signal detection in spontaneous reporting databases, most of these are based on disproportionality methods of case reports.

---

[#] vnikulin.uq@gmail.com

A full safety profile of a new drug can never be known at the time that it is introduced to the general public. Whereas premarketing clinical trials do consider safety endpoints, they are limited in the types and numbers of patients exposed. Actual clinical practice often differs from the controlled setting of a clinical trial, with respect to the indication for treatment, concomitant medication, and dosage at which a drug is prescribed. Also, it may differ over time. As a consequence, safety monitoring and evaluation must continue throughout a drug's life-cycle (Norén et al., 2009).

In this paper we would like to share our successful experience, which was obtained online during the OMOP 2010 Cup. Also, we would like to direct readers to some selected publications (Nikulin, 2008), (Nikulin and McLachlan, 2010) and (Nikulin et al., 2011), where we reported our successful models and methods, which were used during different data mining Challenges.

## OMOP CHALLENGE

OMOP is a public-private partnership designed to improve the monitoring of drugs for safety and effectiveness. The partnership is conducting a two-year research initiative to determine whether it is feasible and useful to use automated healthcare data to identify and evaluate safety issues of drugs on the market. The Partnership's methodological research is conducted across multiple disparate observational databases (administrative claims and electronic health records). The series of studies being conducted include assessing different types of automated healthcare data, developing tools and methods to analyze the databases, and evaluating how analyses can contribute to decision-making.

OMOP relies on the expertise and resources of the U.S. Food and Drug Administration, other federal agencies, academic institutions, the pharmaceutical and health insurance industries and non-profit organizations. A network of institutions, managed by the Foundation for the National Institutes of Health, carries out specific OMOP tasks, and all together, more than 100 partners are collaborating. Throughout the work phases of OMOP all work products are made publicly available to promote transparency and consistency in research.

The competition started in September 2009. OMOP provided the participants with a large simulated data set resembling healthcare data that was "spiked" with adverse events. The competitors had to find the signals by generating methods to identify relationships in the data between drugs and medical outcomes (adverse events). The goal was to develop methods that correctly identified true drug-event associations while minimizing false positive findings. Methods were evaluated by how accurately they predicted the known relationships that existed in the data. At the end of the competition, which was closed on March 31, 2010, there were over sixty competitors from many fields and entities.

*OMOP database*

The given database includes records of 10 million patients with dates when observation was started and ended. The overall observation period is 10 years. For any particular patient we have 2 sequences: 1) drugs with starting and ending dates; 2) conditions with starting date. The total numbers of drugs and conditions are 5,000 and 4,519, respectively. Accordingly, the total number of possible associations is 22,595,000. There are also some demographical information available, such as age and sex. We shall denote by *D* and *C* sets of all drugs and conditions. As an illustration, the organisers made available a small subset of pairs {drug, condition} with true label (4000 positive and 3920 negative), but we did not use this information in the training process.

More details regarding the database and the Challenge may be found on the OMOP website[*]. Most of the pre-processings were conducted using special software written in Perl, the main algorithms were implemented in C. In addition, we used special codes written in Matlab.

## DISPROPORTIONALITY ANALYSIS (DPA)

In studying the temporal association between two events, it is convenient to let one event set the relative time frame in which the incidence of the other event is examined. We shall in the context of this paper let drug prescriptions define the relative time frame in which the incidence of other medical events is examined. The other medical events considered include notes of clinical symptoms, signs, and diagnoses, and prescriptions of other drugs. Our objective is to identify interesting temporal patterns relating the occurrence of a medical event to first prescriptions of a specific drug (Norén et al., 2009).

Let us denote by $\Delta$ a threshold temporal parameter (for example it may be in the range of 30-60 days). Then, we can consider the observation period *T* (for example, it may be the whole period of 10 years). We shall consider all the listed patients and shall compute $n_{dc}$ to be the numbers of associations/cases, where

$$0 \leq t_c - t_d \leq \Delta, \qquad (1)$$

$t_c$ and $t_d$ are the dates when condition *c* and drug *d* were started.

As a next step, we count $n_d$ and $n_c$ to be the numbers of the times drug *d* and condition *c* were found within the time interval *T*. Note that $n_d$ and $n_c$ were computed independently.

Assuming that events are independent, the expected number of associations may be calculated according to the following formula

$$b_{dc} = \lambda_d n_c, \qquad (2)$$

where $\lambda_d = \dfrac{n_d}{N}$, $N = \sum_{d \in D} n_d$.

Finally, we shall compute required ratings

$$r_{dc} = f\left(\dfrac{n_{dc} + \alpha}{b_{dc} + \alpha}\right), \qquad (3)$$

---

[*] *http://omop.fnih.org/omopcup*

where *f* is a logarithmic or power function, $\alpha$ is a smoothing or shrinkage parameter. In our experiments we used $0.1 \leq \alpha \leq 0.5$.

*Mean average precision*

The performance of the solutions was measured using the Mean Average Precision (MAP), metric often used in the field of information retrieval. It measured how well a system ranks items, and emphasizes ranking true positive items higher. It is the average of precisions computed at the point of each of the true positives in the ranked list returned by the method (Schuemie, 2010).

With the approach presented in this section we achieved result *MAP = 0.12* for the Challenge 1.

## TEMPORAL ANALYSIS

Most likely, the ratings (3) will be too rough if they are calculated according to the whole time-interval *T* of 10 years. Therefore, it is proposed to split the whole interval *T* into several consecutive subintervals: $T^{(i)}, i = 1,...,m$, and calculate $r_{dc}^{(i)}, i = 1,...,m$, accordingly. The most suitable value *m = 10*, which corresponds to the number of years within the whole observation period. As an outcome, we can produce solution for Challenge 2 using ratings $r_{dc}^{(i)}, i = 1,...,m$.

*Challenge 2: identifying drug-condition associations as data accumulates over time*

Timely detection of drug-related adverse events as part of an active surveillance system would allow patients and health care providers to minimise potential risks and inform decision-making authorities as quickly as possible. Challenge 2 seeks to evaluate a method's performance in identifying true drug-condition associations and discerning from false association as data accumulates over time.

It is important to mention that an association is defined as a drug that increases the likelihood of a condition occurring. A condition that is less likely to occur after receiving a drug, possibly, as an intentional result of a treatment, is not counted as an association.

For the second challenge, it was necessary to examine the first 500 drugs more closely. As requested, submissions should contain one entry for each such drug-condition combination at the end of each of 10 calendar years, resulting in 10 times 500 times 4519 (22,595,000) total records. That means, the size of all possible combinations for Challenge 2 was exactly the same as for Challenge 1.

We calculated solution for Challenge 2 according to the formula

$$s_{dc}^{(year)} = \frac{1}{year} \sum_{i=1}^{year} r_{dc}^{(i)}, \; year = 1,...,10, \quad (4)$$

and observed *MAP = 0.13*. Also, we considered $s_{dc}^{(10)}$ in application to Challenge 1 with *MAP = 0.14*. Figure 1 illustrates behaviour of the 16 selected (strongest relations) pairs {d, c}, which are presented in Table 1.

## RANDOM RESAMPLING (BAGGING OR HOMOGENEOUS ENSEMBLING)

Bagging predictors is a method for generating multiple versions of a predictor and using these to get an aggregated predictor. The aggregation averages over the versions when predicting a numerical outcome and does a plurality vote when predicting a class (Breiman, 1996). In this section we consider method of random resampling: it is supposed that using the hundreds of predictors (base learners), based on the randomly selected subsets of the whole training set, we shall reduce the random factors. According to the principles of homogeneous ensembling, the final predictor represents an average of the base predictors. As a reference, we mention random forests (Breiman, 2001) is a well-known example of successful homogeneous ensemble. However, the construction of random forests is based on another method, which is linked to the features but not to the samples.

With the method of random resampling we were able to achieve a dramatic improvement in performance: *MAP=0.21* for Challenge 1 and *MAP=0.18* for Challenge 2.

The ratings for Challenge 2 were calculated according to the following formula

$$z_{dc}^{(year)} = \frac{1}{k} \sum_{j=1}^{k} s_{dc}^{(year)}(j), \; year = 1,...,10, \qquad (5)$$

where $j$ is a sequential index of the randomly selected $\Omega_j$ subset of patients, and, by definition, it is assumed that computation of $s_{dc}^{(year)}(j)$ was based on $\Omega_j$.

Table 1: List of 16 strongest (according to our evaluation) relations between drugs and conditions, where ratings were computed according to (4). Column "Figure 1" indicates horizontal label of the window in Figure 1, where this time-series of the corresponding relationship is presented.

| N  | Figure 1 | Drug | Condition | Rating  |
|----|----------|------|-----------|---------|
| 1  | a1       | 198  | 4017      | 6214.97 |
| 2  | a2       | 199  | 4018      | 6105.93 |
| 3  | a3       | 80   | 4011      | 5843.94 |
| 4  | a4       | 3    | 4002      | 5802.4  |
| 5  | b1       | 314  | 4025      | 5700.24 |
| 6  | b2       | 137  | 4013      | 5623.87 |
| 7  | b3       | 362  | 3509      | 5613.11 |
| 8  | b4       | 437  | 4039      | 5585.65 |
| 9  | c1       | 2    | 4002      | 5543.93 |
| 10 | c2       | 471  | 1996      | 5311.32 |
| 11 | c3       | 318  | 4027      | 5302.99 |
| 12 | c4       | 251  | 4020      | 5289.66 |
| 13 | d1       | 79   | 4011      | 5256.8  |
| 14 | d2       | 339  | 4032      | 5233.38 |
| 15 | d3       | 198  | 1280      | 5208.57 |
| 16 | d4       | 3    | 1377      | 5179.46 |

Selection of the patients was conducted according to the condition: $\gamma \leq 0.65$, where $\gamma$ is a standard uniformly distributed random variable. Based on our experiments, the number of random samples $k=100$ is a quite sufficient. In addition, we decided to extend the random sampling further, and used as a threshold parameter $\Delta$ in (1) uniformly distributed random variable: $40 \leq \Delta \leq 60$.

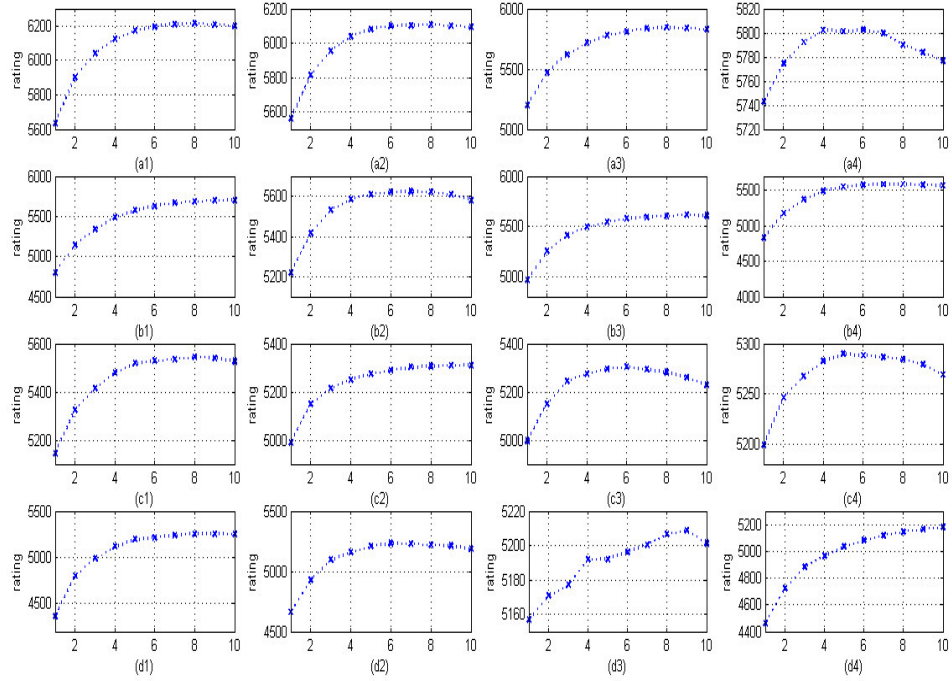

Figure 1. Temporal dependences (ratings as a function of the years) for the selected pairs {drugs, conditions}, which are presented in Table 1.

Note that we used solution $z_{dc}^{(10)}$ for Challenge 1. Figure 2(a) illustrates the structure/histogram of the solution $z_{dc}^{(10)}$, which was reduced to the logarithmic scale, where we used only drugs with indexes from 1 to 500 (this corresponds to Challenge 2). In accordance with Figure 2(a), an absolute majority of the pairs *{d, c}* has no expected relations. Figure 2(b) shows histogram of the right part of the solution presented in Figure 2(a) with some potential links.

## DPA: A SECOND APPROACH BASED ON THE DRUG ERAS

Compared to the first approach of *DPA*, we shall use here not a counter of the number of times when drug was used, but the total duration in days when drug was used.

Let us denote by $h_d$ the total duration of the time when drug *d* was used during observational period *T*. Then, we can rewrite (2) in this way

$$b_{dc} = \theta_d n_c, \qquad (6)$$

where

$$\theta_d = \frac{h_d}{H}, \quad H = \sum_{d \in D} h_d.$$

**Remark 1.** *Based on our experimental evaluations, there is a significant difference between formulas (2) and (6) in terms of the related outcomes. The formulas (2) and (6) are similar in the structural sense, and represent the most important initial steps. The following steps to construct solution for this particular method are the same: we can apply (6) to (3). Then, we can repeat the temporal analysis (4) and resampling (5). As an outcome of this modified procedure we had observed the scores: MAP = 0.225 for Challenge 1 and MAP = 0.205 for Challenge 2.*

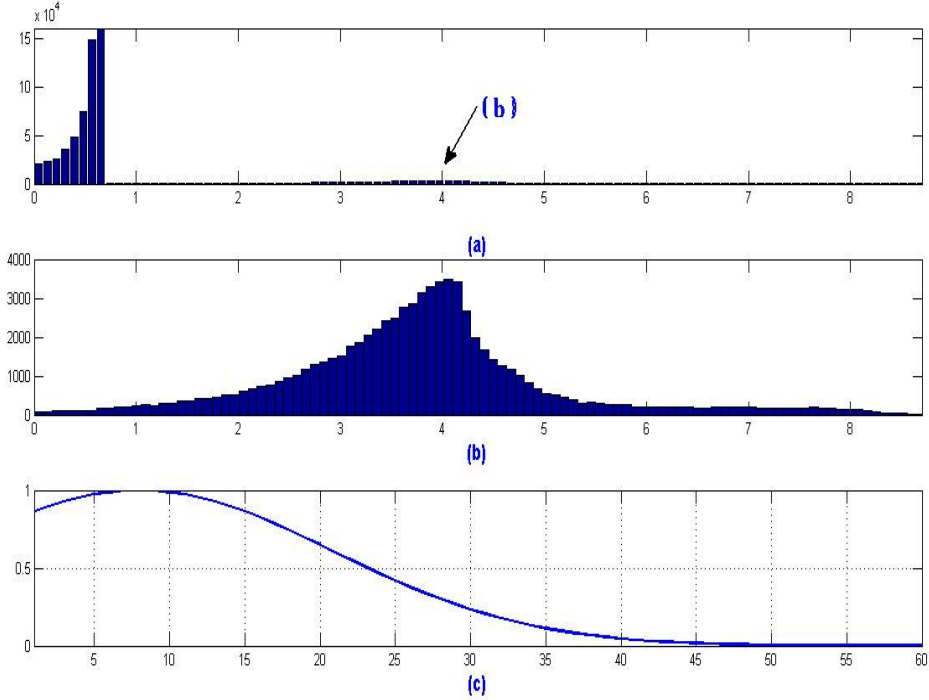

Figure 2. (a) histogram of the solution $z_{dc}^{(10)}$, which is defined in (5); (b) histogram of the right part of the solution $z_{dc}^{(10)}$ (with potential links between drugs and conditions); (c) function $w$ for the temporal weighting.

## HETEROGENEOUS ENSEMBLING

**Definition.** *An ensemble is defined as a heterogeneous if the base models in an ensemble are generated by methodologically different learning algorithms. On the other hand, an ensemble is defined as a homogeneous if the base models are of the same type (for example, resampling or bagging as discussed above).*

As far as solutions *DPA1* and *DPA2*, which are based on expected numbers of associations (2) and (6), are very different in a structural sense, they cannot be linked together directly. At the same time we know that the qualities of both solutions *DPA1* and *DPA2* are high. The later observation represents a very positive factor, which indicates that the solutions *DPA1* and *DPA2* contains different information, which may lead to further improvement if linked in a proper way.

Using an ensemble constructor (Nikulin and McLachlan, 2009), we can adjust one solution to the scale of another solution. After that, we can compute an ensemble solution as a linear combination:

$$ENS = \tau \cdot \overline{DPA1} + (1-\tau) \cdot DPA2, \qquad (7)$$

where $\overline{DPA1}$ is the same as *DPA1* solution, which was adjusted to the scale of *DPA2* solution, $0 < \tau < 1$ is a positive weight coefficient. Clearly, the stronger performance of the solution *DPA2* compared to *DPA1* the smaller will be value of the coefficient $\tau$.

With an ensemble constructor (7), we observed *MAP=0.23* for Challenge 1 and *MAP=0.22* for Challenge 2, where we used $\tau = 0.3$.

## TEMPORAL WEIGHTING FOR COMPUTATION OF THE NUMBERS OF ASSOCIATIONS

According to (3), the value of $n_{dc}$ is a very important. Clearly, the strength of the signal depends essentially on the difference $t_c - t_d$, subject to the condition (1). Based on our statistical analysis (and, also, on some qualitative considerations), we decided to implement the following formula

$$n_{dc} = \sum_{t_d \leq t_c \leq t_d + \Delta} w(t_c - t_d), \qquad (8)$$

where the structure of weight function *w* is illustrated in Figure 2(c): it is logical to assume that reaction of the patient's organism to drug is not an immediate, and the likelihood of the possible association will decline over time after some point (6-10 days).

## COMPUTATION TIME

A Linux multiprocessor computer with speed 3.2GHz, RAM 16GB, was used for the most of the computations. All the algorithms were implemented in C. The running time for 100 random samplings according to (5) was about 10 hours.

## CONCLUDING REMARKS

As a main outcome of our study, we can report very strong improvement with homogeneous ensembling (bagging).

Also, we were trying to differentiate the matrices (5) for the particular age/sex groups, and then create submission assuming that the different age/sex groups are equally important. However, we did not observe any significant improvements with this approach.

During the Challenge we conducted experiments with many different methods and approaches, which were not mentioned in the above Sections. For example, we tried *2D k-means* clustering (Nikulin and McLachlan, 2009) and gradient-based matrix factorisation (Nikulin et al., 2011) in application to the matrix (5) in order to smooth the noise, and can report some modest progress in this direction.

There may be several consecutive eras of one drug for the same patient. We achieved good improvements in the case if we use only first drug era, and ignore all the other eras.

As to the prospective work: assuming that there are true relations for any drug/condition, it maybe a good idea to calibrate the matrix (5) so that the most "shiny" drugs/conditions will not outshine the other drugs/conditions.

According to (Jelizarow et al., 2010), the superiority of new algorithms should always be demonstrated on an independent validation data. In this sense, an importance of the data mining contests is unquestionable. The rapid popularity growth of the data mining challenges demonstrates with confidence that it is the best-known way to evaluate different models and systems.

## ACKNOWLEDGMENTS

We are grateful to the Organisers of the OMOP 2010 data mining Contest for this stimulating opportunity.